\documentclass[letterpaper, 10 pt, conference]{ieeeconf}

\IEEEoverridecommandlockouts                              

\usepackage{graphics} 
\usepackage{epsfig} 

\expandafter\let\csname equation*\endcsname\relax
\expandafter\let\csname endequation*\endcsname\relax
\usepackage{amsmath} 
\usepackage{amssymb}  
\usepackage{siunitx}
\usepackage{nicefrac}
\usepackage{multicol}
\usepackage{multirow}
\usepackage{units}
\usepackage{array}
\usepackage{pgfplots} 
\usepackage{kantlipsum}
\usepackage{setstack}
\pgfplotsset{compat=1.14}
\usepackage{siunitx}
\usepackage{todonotes}

\newcommand{\etal}{\textit{et al.}}

\title{\LARGE \bf
Scaling down an insect-size microrobot, HAMR-VI into HAMR-Jr 
}

\author{Kaushik Jayaram$^{1,2}$, Jennifer Shum$^{1}$, Samantha Castellanos$^{1}$, E. Farrell Helbling$^{1}$ and Robert J. Wood$^{1}$
\thanks{This work is partially funded by the Wyss Institute for Biologically Inspired Engineering, the DARPA SHort Range Independent Microrobotic Platforms (\#HR001119C0051, to R.J.W) and the Department of Defense (DoD) through the National Defense Science \& Engineering Graduate (NDSEG) Fellowship Program (to J.S.). The views, opinions,  and/or findings expressed here, are those of the authors and should not be interpreted as representing the official views or policies of the Department of Defense or the U.S. Government.}
\thanks{$^{1}$Harvard Microrobotics Lab, School of Engineering and Applied Sciences, Harvard University and Wyss Institute for Biologically Inspired Engineering
        {\tt\small \{kjayaram, rjwood\}@seas.harvard.edu}}%
\thanks{$^{2}$Paul M. Rady Department of Mechanical Engineering, University of Colorado Boulder, 
        {\tt\small kaushik.jayaram@colorado.edu}}%
}

\begin{document}

\maketitle
\thispagestyle{empty}
\pagestyle{empty}

\begin{abstract}

 Here we present HAMR-Jr, a \SI{22.5}{\milli\meter}, \SI{320}{\milli\gram} quadrupedal microrobot. With eight independently actuated degrees of freedom, HAMR-Jr is, to our knowledge, the most mechanically dexterous legged robot at its scale and is capable of high-speed locomotion (\SI{13.91}{bodylengths~\second^{-1}}) at a variety of stride frequencies (\SI{1}{}-\SI{200}{\hertz}) using multiple gaits. We achieved this using a design and fabrication process that is flexible, allowing scaling with minimum changes to our workflow. We further characterized HAMR-Jr's open-loop locomotion and compared it with the larger scale HAMR-VI microrobot to demonstrate the effectiveness of scaling laws in predicting running performance.

\end{abstract}

\section{INTRODUCTION}

In recent years, there has been a growing interest in micro-to-milli (i.e., insect scale) robotics research \cite{diller2013micro}\cite{st2019toward}. These robots have the potential for tremendous social and economic impact in the areas of search-and-rescue, high value asset inspection, environmental monitoring, and even medicine. Successful examples of such robots have been discussed in a recent review by St. Pierre \etal, \cite{st2019toward}. 

Advances in micro and meso scale fabrication have enabled new technologies for constructing and operating miniature robots, such as rapid prototyping, additive fabrication, high bandwidth actuators, and low power sensors. However, a majority of existing insect-scale robots tend to have simple body plans, possess limited intelligence, and lack power and control autonomy beyond basic locomotion. Some recent scientific efforts are beginning to address these challenges (e.g., untethered flight \cite{jafferis2019untethered} and multimodal robotic collectives \cite{zhakypov2019designing}).

One of the smallest and fastest legged robots at this scale is the Harvard Ambulatory MicroRobot (HAMR), a quadrupedal microrobot. Early design and manufacturing of HAMR was explored by Baisch \etal, and HAMR was shown to run at speeds above \SI{44}{\centi\meter\second^{-1}} – corresponding to \SI{10}{bodylengths~\second^{-1}} ($BLs^{-1}$) using multiple gaits. Recent efforts have focused on adding multi-modal locomotion capabilities \cite{deRivazeaau3038}\cite{chen_controllable_2018}, power autonomy \cite{goldberg2018power}, sensing \cite{jayaram2018concomitant} and closed-loop control \cite{doshi2017phase}\cite{doshi2019effective}, and optimization based gait generation \cite{doshi2018contact}. 

\begin{figure}[t]
	\begin{center}
		{\includegraphics[width=\columnwidth]{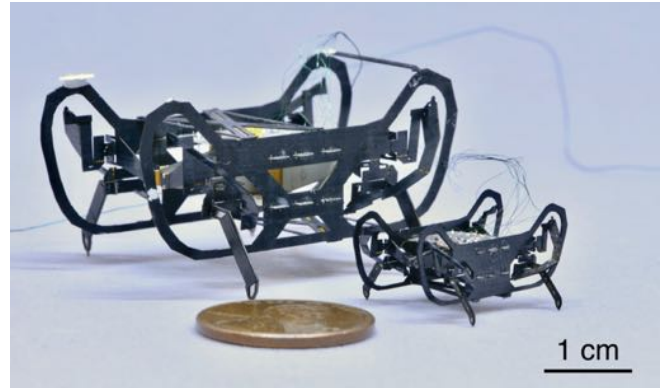}}
		\vspace{-0.5cm}
		\caption{HAMR-Jr (right) alongside its predecessor, HAMR-VI (left). HAMR-Jr is only slightly bigger in length and width than a penny (19 mm diameter) measuring 22.5 mm in body length and weighs a mere \SI{320}{\milli\gram} making it one of the smallest yet highly capable, high-speed insect-scale robots.}
		\label{fig:hamrjr}
		\vspace{-0.5cm}
	\end{center}
\end{figure}

In this work, we introduce a new smaller-scale version of HAMR-VI, hereafter referred to as HAMR-Jr (Figure \ref{fig:hamrjr}). This robot builds on extensive previous work on the HAMR platform \cite{baisch2014high,doshi2015model,goldberg2017gait,doshi2019effective}. In particular, its design is derived from HAMR-VI (\SI{4.51}{\centi\meter} long, \SI{1.43}{\gram} mass,  \cite{doshi2015model}), a robot capable of achieving high-speed, level ground locomotion using multiple closed-loop gaits \cite{doshi2019effective} at a variety of leg cycling frequencies \cite{goldberg2017gait}. Despite measuring half the body length and less than one fourth the body mass relative to its larger counterpart, HAMR-Jr does not compromise any mechanical dexterity or control authority (Figure \ref{fig:hamrjr}). 

HAMR-Jr made possible by the PC-MEMS fabrication process \cite{sreetharan2012monolithic} and pop-up assembly techniques \cite{whitney2011pop}, which allow for fast and repeatable assembly of the complex leg transmissions, as seen in Figure \ref{fig:hamr_transmission}. This scale-independent design and fabrication workflow allows us to investigate a number of at-scale hypotheses (e.g., template dynamics for running). Here, we demonstrate this methodology with a half-scale version. While a similarly-sized robot was previously developed \cite{baisch2013pop} to demonstrate a proof of concept of the strength of our pop-up fabrication technique, no systematic characterization was performed. Further, as it was derived from an older generation (HAMR-VP) designed with only six DOFs, it had limited locomotion capabilities.

The main contributions of the paper are organized as follows. We describe the design and fabrication of HAMR-Jr in section \ref{sec:robot_design}. Further, we discuss the scaling of actuators, flexure joints, and chassis that enable miniaturization and predict its effect on locomotion performance. In section \ref{sec:robot_perf}, we characterize the resulting transmission dynamics and quantify the key performance metrics of the robot kinematics (speed, payload, cost of transport) during level running using two gaits (trot and pronk). 
We conclude by discussing the validity of our scaling hypotheses and identify potential directions for future work in sections \ref{sec:discussion} and \ref{sec:future}, respectively.

\section{ROBOT DESIGN AND CHARACTERIZATION}
\label{sec:robot_design}
\subsection{Platform Overview}

\begin{figure}[ht]
	\begin{center}
		{
		\includegraphics[width=\columnwidth]{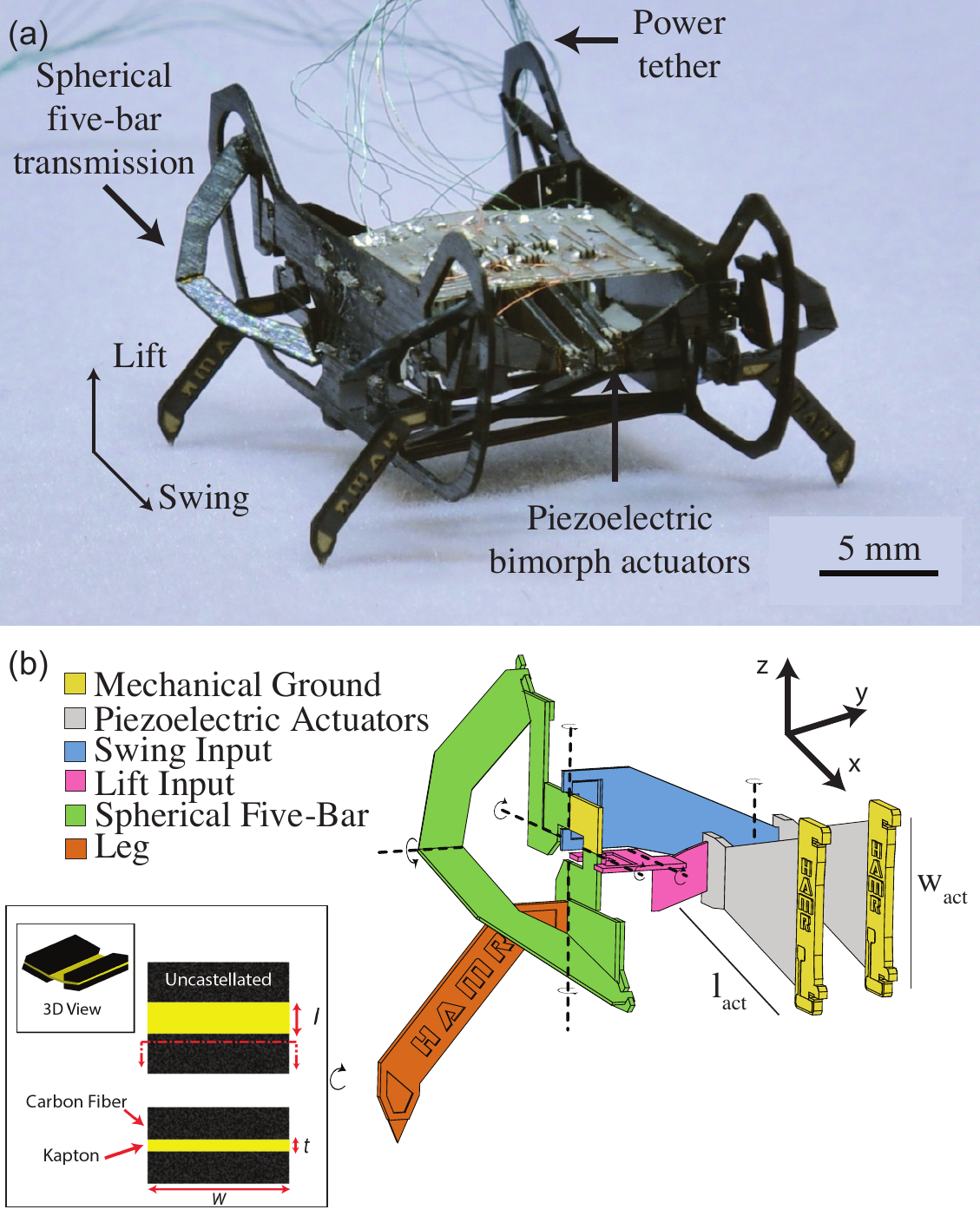}
		}
		\vspace{-0.5cm}
		\caption{(a) A perspective view of HAMR-Jr with components labeled. Also indicated are the axes of motion for lift and swing DOFs. (b) Schematic of HAMR-Jr's transmission (rear right leg) with subsystems color coded. The length and width of the actuators are also denoted. Inset on the bottom left is a schematic of the flexure joints highlighting the relevant dimensions.}
		\label{fig:hamr_transmission}
		\vspace{-0.5cm}
	\end{center}
\end{figure}

HAMR-Jr, as depicted in Figure \ref{fig:hamr_transmission}a, is a \SI{2.25}{\centi\meter} long, \SI{320}{\milli\gram} quadrupedal microrobot with eight independently actuated degrees of freedom (DOFs) driven by high-bandwidth, self-sensing \cite{jayaram2018concomitant} piezoelectric actuators \cite{jafferis2015design} and capable of achieving unprecedented locomotion frequencies of over \SI{200}{\hertz}. 

A schematic of a single leg with labeled components is depicted in Figure \ref{fig:hamr_transmission}b (adapted from \cite{doshi2015model}). Each leg has two active DOFs, a lift DOF that controls leg motion in the YZ-plane and a swing DOF that controls leg motion in the XY-plane. The use of optimal energy density piezoelectric bending actuators \cite{jafferis2015design} for both DOFs allows HAMR-Jr to operate at a wide range of frequencies (\SI{1}{}-\SI{280}{\hertz}). 

\subsection{Scaling of Actuators, Flexures, and Chassis}
\label{sec:scaling}
Due to material availability constraints, we were unable to shrink HAMR-VI into HAMR-Jr isometrically. Therefore, we chose to scale down allometrically -- we maintained the thickness of the robot components (chassis and actuators) but reduced the length ($l$) and width ($w$) by half. As a result, we hypothesize the following physical dimensions for the miniaturized robot chassis:
\begin{align}
\label{eqn:scale_chassis}
    l_{jr} = \frac{l_{vi}}{2} \nonumber\\ 
    w_{jr} = \frac{w_{vi}}{2} \nonumber\\
    BL \propto l \implies BL_{jr} = \frac{BL_{vi}}{2}\nonumber\\
    m \propto lw \implies m_{jr} = \frac{m_{vi}}{4},
\end{align}
where $BL$ is the body length of the robot, $m$ is the body mass and the subscripts $jr$ and $vi$ indicate the scaled (HAMR-Jr) and original-sized (HAMR-VI) robot prototypes.

Similarly, from our previous work \cite{jafferis2016multilayer}, we expect the actuator mechanical characteristics to vary as:
\begin{align}
\label{eqn:scale_actuator_mech}
    d \propto l_{act}^2 \implies d_{jr} = \frac{d_{vi}}{4} \nonumber\\
    F \propto \frac{w_{act}}{l_{act}} \implies F_{jr} = F_{vi} \nonumber\\
    k^{act} \propto \frac{w}{l^3} \implies k^{act}_{jr} = 4k^{act}_{vi},
\end{align}
where $d$ is the tip deflection, $F$ is the blocked force and $k$ is the actuator stiffness. Further, the electrical properties of the actuator \cite{jayaram2018concomitant} scale as: 
\begin{align}
\label{eqn:scale_actuator_dielec}
    C \propto lw \implies C_{jr} = \frac{C_{vi}}{4} \nonumber\\
    R \propto \frac{1}{lw} \implies R_{jr} = 4R_{vi},
\end{align}
where $C$ and $R$ are the voltage- and frequency-dependant actuator capacitance and resistance, respectively.

Additionally, we want the flexure joints to maintain the same range of motion. We scaled the width of the flexure, while keeping the length constant. Therefore, the mechanical properties \cite{doshi2015model} vary as: 
\begin{align}
\label{eqn:scale_flexure}
    \theta \propto l \text{ (constant)} \implies \theta_{jr} = \theta_{vi} \nonumber\\
    k^{flex} \propto \frac{w}{l} \implies k^{flex}_{jr} = \frac{k^{flex}_{vi}}{2},
\end{align}
where $\theta$ is the maximum angular deflection and $k^{flex}$ is the stiffness of the joints.
Since the actuators and joint flexures act in parallel in our transmission (
Figure \ref{fig:hamr_transmission}b, \cite{doshi2015model}), we hypothesize the overall transmission stiffness ($k^{tot}$) to scale as:
\begin{align}
\label{eqn:scale_stiffness}
    k^{tot} = k^{act} + k^{flex} \implies k^{tot}_{jr} = \frac{9k^{tot}_{vi}}{2}.
\end{align}
As a result, we expect the natural resonant frequencies ($\omega$) of our miniaturized robot to scale as: 
\begin{align}
\label{eqn:scale_resonance}
     \omega = \sqrt{\frac{k}{m}} \implies \omega_{jr} = \sqrt{18}\omega_{vi} \approx 4.24\omega_{vi}.
\end{align}
Finally, since HAMR-Jr and HAMR-VI have the same transmission ratios between the actuator and leg output, we can use the actuator tip deflection (equation \ref{eqn:scale_actuator_mech}) as a proxy for stride length. Similarly, we can utilize the resonant frequencies (equation \ref{eqn:scale_resonance}) to define the bandwidth of leg cycling frequency. Assuming that robots at both sizes rely on similar template dynamics \cite{dickinson2000animals}, we can predict the effect of scaling on maximal performance (running speed) as:
\begin{align}
\label{eqn:scale_performance}
     v \propto d\omega \implies v_{jr} = \frac{\sqrt{18}}{4}v_{vi} \approx 1.06v_{vi},
\end{align}
where $v$ is the mean running speed. We thus hypothesize that the maximal locomotion speed for the miniaturized HAMR-Jr should be similar to that of HAMR-VI in absolute units (\SI{}{\milli\meter\second^{-1}}) and about twice as fast in body length normalized units (\SI{}{BL\second^{-1}}). 

\subsection{Transmission Dynamics}
\label{sec:trans_dyn}
In order to verify that the above scaling relations were realized in HAMR-Jr, we experimentally characterized the transmission, as described below. 
\subsubsection{Vertical Leg Stiffness}
\label{sec:trans_stiff}
Template dynamics resulting from bioinspired locomotion models \cite{dickinson2000animals} such as Single Leg Inverted Pendulum (SLIP, \cite{cavagna1977mechanical}) and Lateral Leg Spring (LLS, \cite{full1999templates}) have informed the design of robotic systems including HAMR. Specifically, recent studies \cite{doshi2019effective} on HAMR have demonstrated the importance of vertical leg stiffness to robot locomotion performance as a function of locomotion frequency. 

\begin{figure}[ht]
	\begin{center}
		{
		\includegraphics[width=\columnwidth]{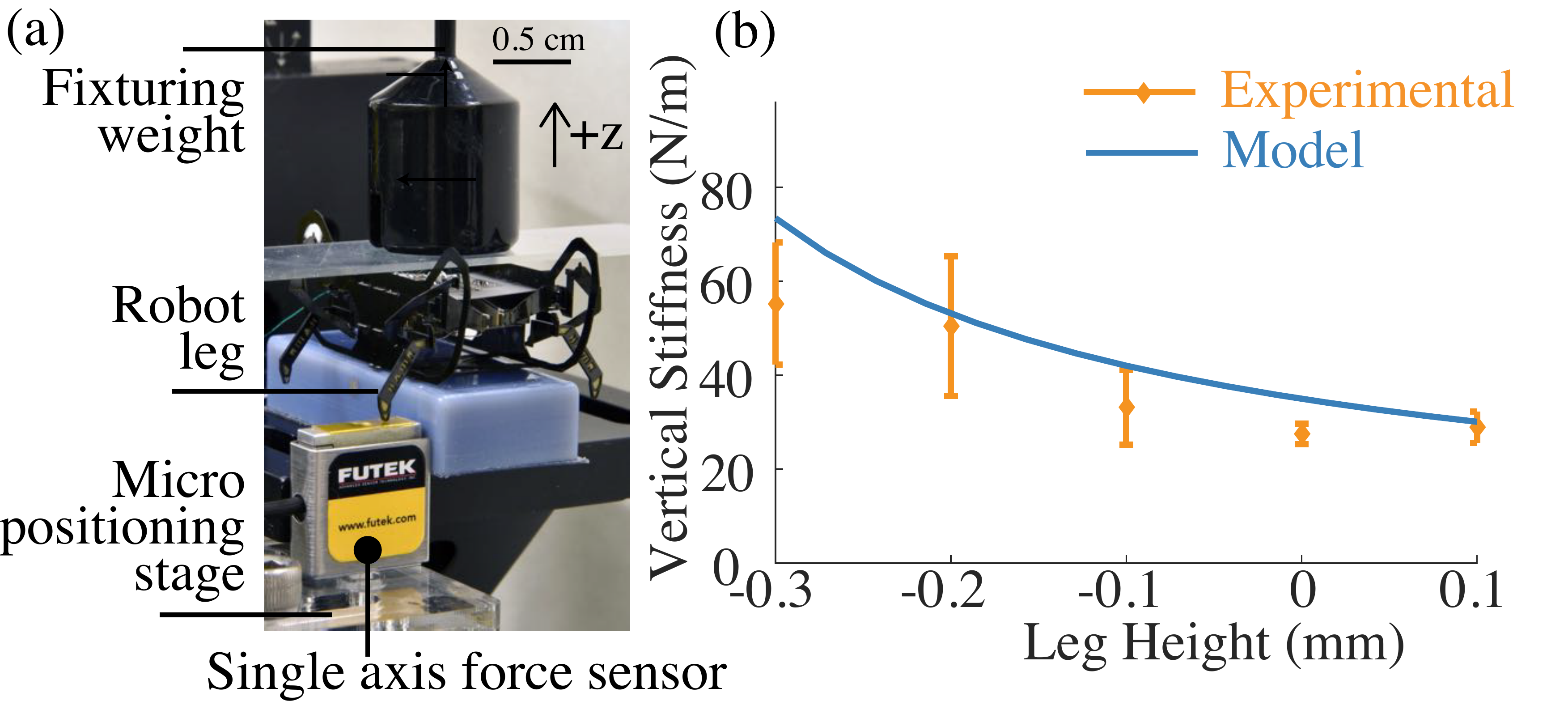}
		}
		\vspace{-0.5cm}
		\caption{(a) Experimental setup used to determine vertical leg stiffness of HAMR-Jr (b) Vertical leg stiffness is plotted as a function of leg height. The experimentally measured data (mean$\pm$1s.d.) is plotted in tan. The model estimates \cite{doshi2015model} are plotted as a blue trace.}
		\label{fig:trans_stiff}
		\vspace{-0.25cm}
	\end{center}
\end{figure}

As a first step towards predicting the sagittal plane dynamics resulting from scaling down, we experimentally determined the stiffness of the lift transmission using the procedure described in Doshi \etal, \cite{doshi2019effective}. Using the experimental setup depicted in Figure \ref{fig:trans_stiff}a, we measured vertical stiffness as a function of leg height for each of the four lift transmissions and found that it ranged from \SI{34.52}{} to \SI{72.11}{\newton\meter^{-1}} (Figure \ref{fig:trans_stiff}b). This matches the same trend as the model \cite{doshi2015model}\cite{doshi2018contact} predictions where the transmission is softest at the highest leg position and becomes more than 2$\times$ stiffer at the lowest tested position \cite{doshi2019effective}. Furthermore, we find that these values are roughly four times that for the same transmission on the larger HAMR-VI \cite{doshi2019effective}, supporting our scaling hypothesis (Equation \ref{eqn:scale_stiffness}).

\subsubsection{Frequency Response}
\label{sec:trans_freq_resp}
To develop an empirical understanding of HAMR-Jr's transmission dynamics, we measured its resonance properties in air by driving the lift and swing transmissions individually (with the other held at its nominal position) by a \SI{40}{\volt}, \SI{16}{\second} long linear chirp signal between \SI{1}{}-\SI{400}{\hertz}. We followed the experimental procedure and used the experimental setup described in detail by Doshi \etal, \cite{doshi2017phase} to test each of the four swing and lift transmissions.

\begin{figure}[ht]
	\begin{center}
		{
		\includegraphics[width=\columnwidth]{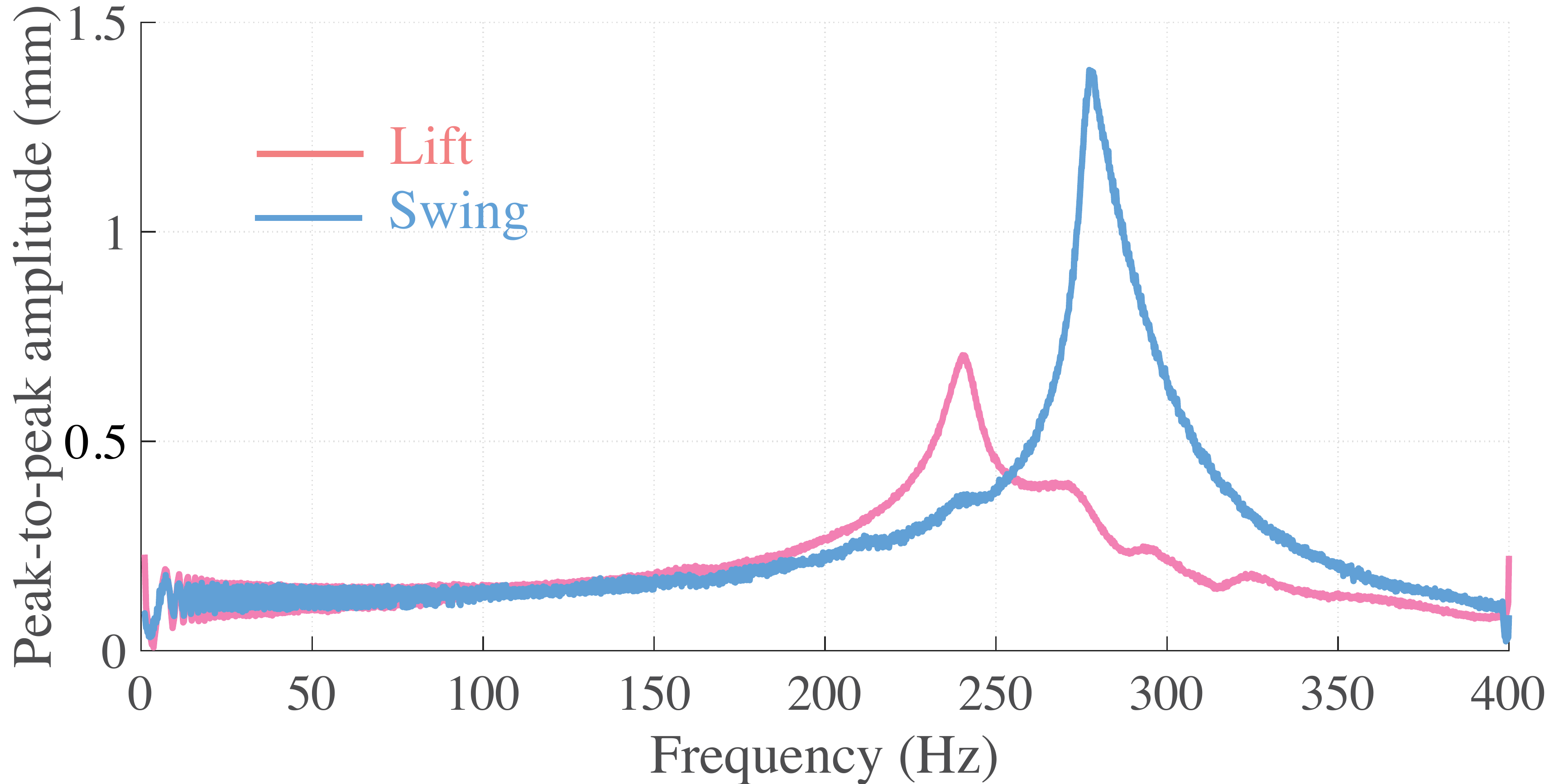}
		}
		\vspace{-0.5cm}
		\caption{Frequency response of the HAMR-Jr transmissions. Plotted is the peak-to-peak amplitude (\SI{}{\milli\meter}) as a function of the drive frequency (\SI{}{\hertz}) for a \SI{40}{\volt} linear chirp input for lift (pink) and swing (blue) transmissions, respectively.}
		\label{fig:freq_resp}
		\vspace{-0.5cm}
	\end{center}
\end{figure}

Figure \ref{fig:freq_resp} indicates that the two transmission DOFs have separate resonances, with the lift resonance at \SI{237.3}{}$\pm$\SI{6.8}{\hertz} (pink trace), and the swing resonance at \SI{279.1}{}$\pm$\SI{5.3}{\hertz} (blue trace). The location of these peaks occurs at much higher frequencies relative to those of the larger HAMR-VI (\SI{81.3}{}$\pm$\SI{3.8}{\hertz} for lift, \SI{103.0}{}$\pm$\SI{3.3}{\hertz} for swing, \cite{doshi2017phase}), albeit less than what is predicted ($\approx$\SI{345}{\hertz} for lift, $\approx$\SI{437}{\hertz} for swing) by the scaling laws (equation \ref{eqn:scale_resonance}). Both the resonant peaks of the HAMR-Jr transmission have a high quality factor, with $Q_{lift}$ = \SI{ 6.3}{}$\pm$\SI{3.3}{} for the lift, and $Q_{swing}$ = \SI{9.6}{}$\pm$\SI{1.5}{} for the swing. This indicates a potentially significant increase in vertical oscillations at the lift resonance and stride length at the swing resonance, if we can operate under those conditions. 

\section{ROBOT PERFORMANCE}
\label{sec:robot_perf}
\subsection{Locomotion Gaits}
\label{sec:gaits}
HAMR-Jr can perform a variety of locomotion gaits by modulating the relative phase between it's eight independently-actuated DOFs. The supplementary video demonstrates four commonly observed bioinspired gaits; trot, pronk, bound, and jump, as defined by their footfall patterns described in Goldberg \etal, \cite{goldberg2017gait}. The robot is also able walk sideways (crab-like movement) and turn using open-loop commands.   

\subsection{Level Running Kinematics}
\label{sec:kinematics}
In order to quantify the kinematics during open-loop level ground running, we chose to experimentally quantify running speed and stride length as a function of stride frequency (eleven discrete values ranging from \SI{1}{} to \SI{280}{\hertz}) for trot and pronk gaits at a single actuator drive voltage of \SI{200}{\volt}. Although the actuators can be driven at frequencies exceeding 280 Hz, transmission resonances (described in section \ref{sec:trans_freq_resp}) limited locomotion at higher frequencies. We performed five trials at each operating condition. 

\begin{figure}[ht]
	\begin{center}
		{
		\includegraphics[width=\columnwidth]{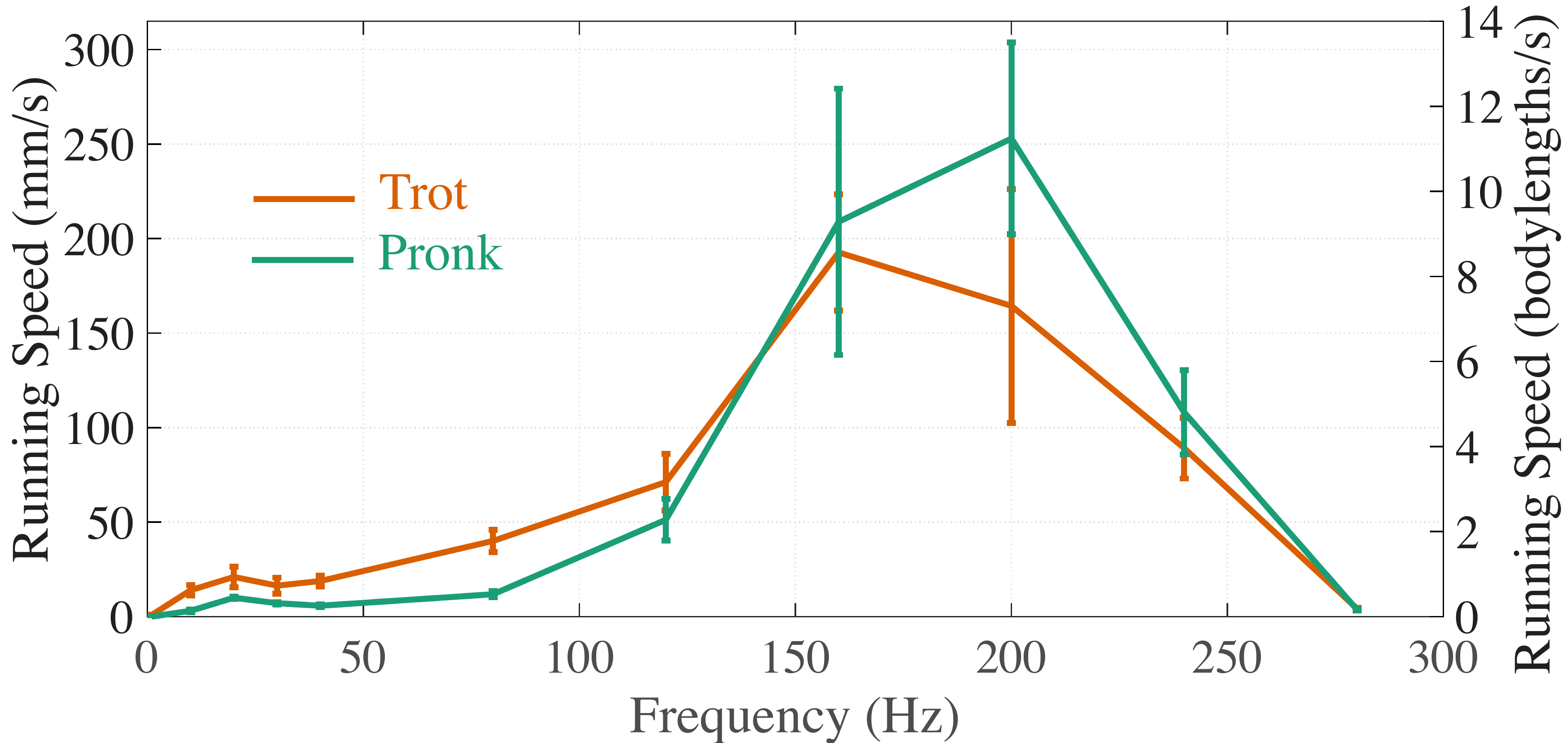}
		}
		\vspace{-0.25cm}
		\caption{Plot of running speed as a function of frequency for HAMR-Jr during locomotion using trot (tan) and pronk (green) gaits. The data represents mean$\pm$1s.d.}
		\label{fig:speed_char}
		\vspace{-0.5cm}
	\end{center}
\end{figure}

While trotting, we measured that HAMR-Jr exhibited average running speeds ranging from \SI{0.91}{}-\SI{192.68}{\milli\meter\sec^{-1}} as depicted in Figure \ref{fig:speed_char} (tan trace). These correspond to \SI{0.04}{}-\SI{8.56}{BL\sec^{-1}} in body size normalized locomotion speeds and matches the performance measured in HAMR-VI \cite{goldberg2017gait}\cite{doshi2019effective} and other HAMR platforms \cite{baisch2014high}. As expected, running speed increases as a function of stride frequency until we approach the lift and swing transmission resonances around \SI{240}{\hertz} and \SI{280}{\hertz}, respectively. The highest measured locomotion speed was \SI{259.49}{\milli\meter\sec^{-1}} (\SI{11.54}{BL\sec^{-1}}) at a \SI{160}{\hertz} stride frequency (supplementary video).

While pronking, we observed that HAMR-Jr's locomotion performance is comparable to that during trotting and, we measured average running speeds ranging from \SI{0.05}{}-\SI{252.95}{\milli\meter\sec^{-1}} (\SI{0.01}{}-\SI{11.24}{BL\sec^{-1}}, Figure \ref{fig:speed_char}, green trace). These results again compare favorably against the corresponding data from previously published research on the HAMR-VI \cite{goldberg2017gait}\cite{doshi2019effective}. The highest measured locomotion speed was \SI{313.03}{\milli\meter\sec^{-1}} (\SI{13.91}{BL\sec^{-1}}) at a \SI{200}{\hertz} stride frequency (supplementary video).

\begin{figure}[ht]
	\begin{center}
		{
		\includegraphics[width=\columnwidth]{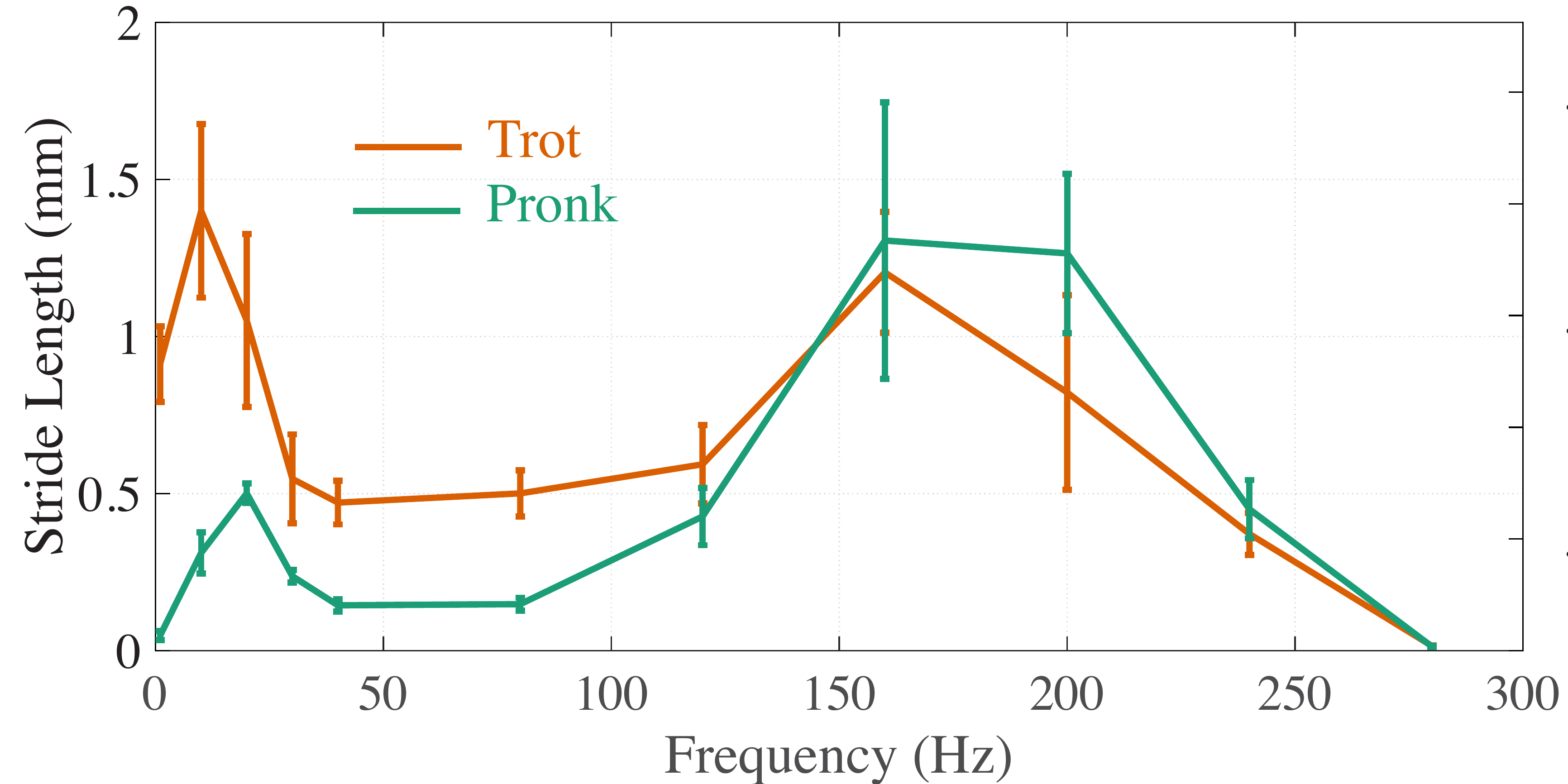}
		}
		\vspace{-0.5cm}
				\caption{Plot of stride length as a function of frequency for HAMR-Jr during locomotion using trot (tan) and pronk (green) gaits. The data represents mean$\pm$1s.d.}
		\label{fig:strlen_char}
		\vspace{-0.5cm}
	\end{center}
\end{figure}

To further characterize the robot's kinematic performance, we quantified the mean effective stride length during locomotion (Figure \ref{fig:strlen_char}). Based on transmission kinematics and assuming no slip on the ground, the theoretical estimate of the stride length for trot and pronk gaits during quasi-static locomotion is \SI{2.08}{\milli\meter} and \SI{1.04}{\milli\meter}, respectively. We measured effective stride lengths ranging from \SI{0.13}{}-\SI{1.77}{\milli\meter} while trotting and \SI{0.01}{}-\SI{1.31}{\milli\meter} while pronking. These are less than the theoretical estimates, indicating that the robot slips considerably. In particular, we found that the robot's stride lengths significantly exceeded the quasi-static estimates during locomotion at the highest running speeds (\SI{160}{}-\SI{200}{\hertz}), despite instances of foot slippage. This is a clear indication that the robot is taking advantage of favorable SLIP-like dynamics for achieving faster locomotion. While this might be applicable in other frequency regimes, we suspect this effect is less pronounced due to increased foot slipping and unfavorable body oscillations resulting from poor ground contact.     


\subsection{Payload}
\label{sec:payload}
To quantify the payload capacity of HAMR-Jr, we added custom 3-D printed weights on top of the robot and ran it for five trials, each at \SI{10}{\hertz}, using trot and pronk gaits. We found that the robot's performance was not significantly affected with an additional payload equal to or less than the robot's body mass. However, larger payloads forced the robot to slow down. This performance is worse than its counterpart, HAMR-VI, which can carry up to $2\times$ its body weight before significant speed reduction \cite{doshi2015model}. This result is surprising and we suspect that stress softening of flexure joints from fatigue, coupled with unmodeled off-axis compliance in the transmission are potential causes for the low payload capacity.

\subsection{Cost of Transport}
\label{sec:cot}
\begin{figure}[ht]
	\begin{center}
		{
;                    		\includegraphics[width=\columnwidth]{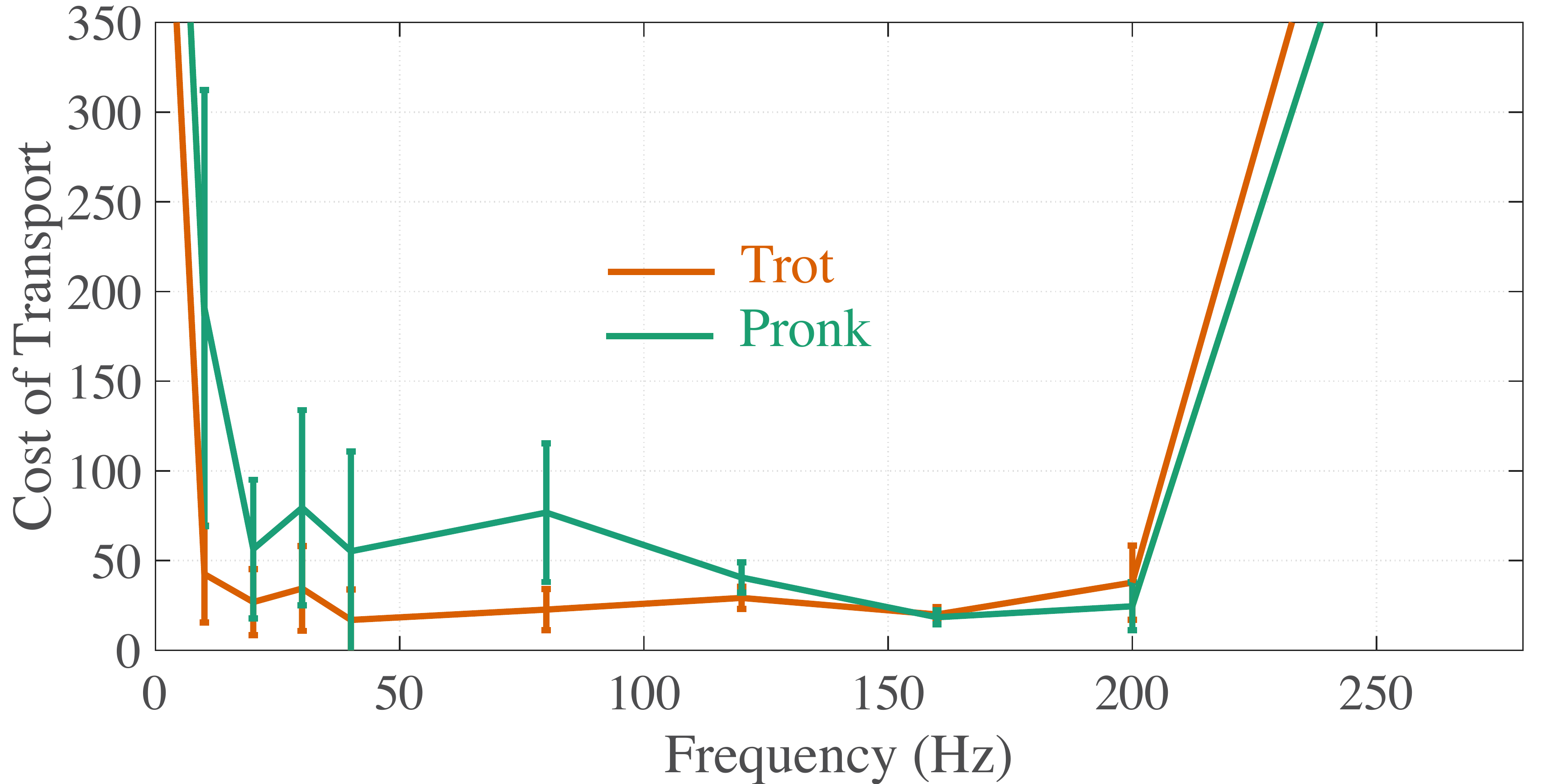}
		}
		\vspace{-0.5cm}
				\caption{Plot of cost of transport as a function of frequency for HAMR-Jr during locomotion using trot (tan) and pronk (green) gaits. The data represents mean$\pm$1s.d.}
		\label{fig:cot_char}
		\vspace{-0.5cm}
	\end{center}
\end{figure}

Cost of transport ($CoT$) is a measure of the energy consumption \cite{dickinson2000animals} during locomotion and often a key parameter critical to extending the operational time of an autonomous robot. We quantified CoT as:
\begin{align}
\label{eqn:cot}
     CoT = \frac{\sum_{j=1}^{8}\frac{1}{T} \int_{0}^{T}i_j(t)V_j(t) dt}{mgv},
\end{align}
where $i_j(t)$ and $V_j(t)$ is instantaneous current and voltage of the $j^{th}$ actuator (typically \SI{10}{}-\SI{100}{\micro\ampere} at \SI{200}{\volt}), and $T$ is stride period. We performed five trials at each frequency for each gait using the approach detailed by Doshi \etal \cite{doshi2019effective}. We found that $CoT$ ranged from \SI{33.8}{} to \SI{9716}{} (Figure \ref{fig:cot_char}). In particular, $CoT$ was high (\SI{>100}{}) at the extremes of the stride frequency range independent of gait due to slow (\SI{1}{\hertz}) or ineffective locomotion (\SI{>200}{\hertz}, approaching transmission resonances). However, for the majority of stride frequencies (\SI{10}{\hertz}-\SI{200}{\hertz}), $CoT$ was typically less than 100 and in the same range as that observed for HAMR-VI \cite{goldberg2017gait}\cite{doshi2019effective}. Furthermore, we found that $CoTs$ are typically lower for locomotion with the trot gait compared to the pronk across the range of stride frequencies. These suggest that the robot is able to utilize more favorable dynamics (SLIP-like), as well as energy storage and return mechanisms while trotting.

\subsection{Proprioceptive Sensing}

\begin{figure}[ht]
	\begin{center}
		{
		\includegraphics[width=\columnwidth]{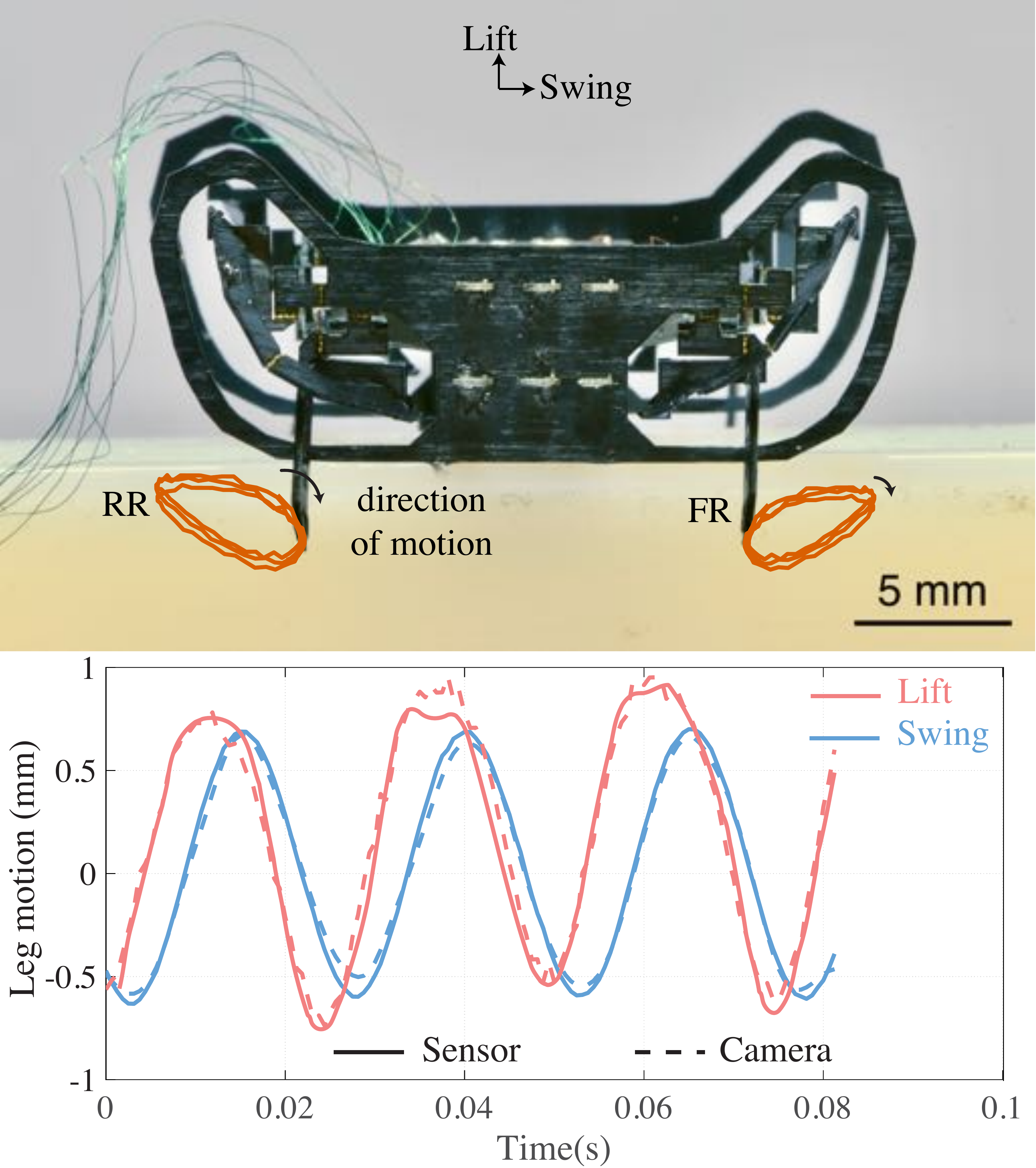}
		}
		\vspace{-0.5cm}
		\caption{(a) Side view of HAMR-Jr, with typical foot trajectories plotted while trotting at \SI{160}{\hertz} and \SI{200}{\volt}. (b) Plot of leg motion as function of time when driving the robot at \SI{160}{\hertz} and \SI{200}{\volt} in air. Motion in lift and swing DOFs are drawn as pink and blue traces respectively. The sensor data is shown as solid traces, while the tracking data from high-speed videography is shown as dashed traces.}
		\label{fig:foot_track}
		\vspace{-0.5cm}
	\end{center}
\end{figure}

Here, we leveraged recent research findings on concomitant sensing and actuation of piezoelectric actuators \cite{jayaram2018concomitant} and its successful application for proprioceptive sensing and closed-loop control during locomotion with HAMR-VI \cite{doshi2019effective}\cite{doshi2018contact}. This technique relies on the insight that motion of the actuators causes varying strains on the surface on the piezoelectric material, which via the direct piezoelectric effect, results in a current (measured via off-board encoders) proportional to the actuator velocity. Using the scaled estimates of the actuator electrical properties (equation \ref{eqn:scale_actuator_dielec}), we followed the procedure described by Doshi \etal, \cite{doshi2019effective} to successfully estimate the position of the feet and verify this motion using a high-speed video camera (Figure \ref{fig:foot_track}). 

\section{IMPLICATIONS OF SCALING DOWN}
\label{sec:discussion}

By successfully fabricating and operating HAMR-Jr, we have demonstrated the ability to scale down HAMR-VI, already one of the smallest, fastest, and most mechanically dexterous robots. Table \ref{tab:hamr_comp} provides an overview of the physical dimensions, mechanical properties, and locomotion performance of HAMR-Jr and HAMR-VI. We also compare HAMR-Jr's locomotion metrics against a few recent insect-scale robots (body length \SI{<5}{\centi\meter}, body mass \SI{<2}{\gram}) in Table \ref{tab:hamr_other_comp}. Similar to robots at this scale, HAMR-Jr is able to operate at stride frequencies higher than those observed in the biological organisms. However, the distinguishing characteristic of HAMR-Jr is the robot's ability to achieve high locomotion speeds across its frequency range while using multiple gaits.   

\begin{table}[ht]
\caption{Comparison of the key characteristics of HAMR-Jr against HAMR-VI}
\label{tab:hamr_comp}
\begin{center}
\begin{tabular}{l|c|c|c}
\hline
\textbf{Robot} & \textbf{HAMR-Jr} & \textbf{HAMR-VI} & \textbf{Relative}\\
& & & \textit{Scaling}\\
& & & \textit{Experimental}\\
& & & \textit{(Theoretical)}\\
\hline
\hline
\multicolumn{3}{l}{\textbf{\textit{Physical dimensions}}}\\
\hline
Body Length (\SI{}{\milli\meter}) & 22.5 & 45.1 & 0.5 (0.5)\\
Body Mass (\SI{}{\gram}) & 0.32 & 1.41 & 0.23 (0.25)\\
\hline
\multicolumn{3}{l}{\textbf{\textit{Mechanical properties}}}\\
\hline
Vertical Stiffness & 32.42 & 9.21 & 3.52 (4.5)\\
Lift Resonance (\SI{}{\hertz}) & 237.3 & 81.3 & 2.91 (4.24)\\
Swing Resonance (\SI{}{\hertz}) & 279.1 & 102.3 & 2.73 (4.24)\\
\hline
\multicolumn{3}{l}{\textbf{\textit{Maximal Locomotion Performance}}}\\
\hline
Stride Frequency (\SI{}{\hertz}) & 200 & 65 & 3.07 (4.24)\\
Stride Length (\SI{}{\milli\meter}) & 1.9 & 8.4 & 0.23 (0.25)\\
(Quasi-static at \SI{200}{\volt}) & & &\\
Stride Length (\SI{}{\milli\meter}) & 1.77 & 10.87 & 0.17 (0.25)\\
Speed (\SI{}{\milli\meter\second^{-1}}) & 313 & 478 & 0.66 (1.06)\\
Speed (\SI{}{BL\second^{-1}}) & 13.9 & 10.6 & 1.31 (2.12)\\
CoT (minimum) & 33.8 & 7.4 & - \\
\hline
\end{tabular}
\end{center}
\end{table}

In summary, we found that HAMR-Jr behaved as predicted, for the most part matching our performance expectations and successfully validating our scaling hypotheses described in section \ref{sec:scaling}. Like its larger predecessors, HAMR-Jr can run at a range of stride frequencies using various gaits. Furthermore, HAMR-Jr has recorded the fastest speed normalized to body size amongst the HAMR series of robots. In fact, this is faster than most other legged insect-scale robots \cite{diller2013micro}\cite{st2019toward} and second only to the ultralight soft robot recently developed by Wu \etal, \cite{wu2019insect} as highlighted in Table \ref{tab:hamr_other_comp}. Furthermore, this locomotion performance compares favorably to that of similarly-sized invertebrates \cite{full2010invertebrate} and begins to approach some of the running speeds of the fastest-recorded insect runners \cite{full1991mechanics}\cite{merritt1999fastest}\cite{rubin2016exceptional}. Despite these impressive results, HAMR-Jr's running speeds are less than our scaling analysis predicted (equation \ref{eqn:scale_performance}) based on HAMR-VI performance. High-speed videography of the robot's locomotion revealed multiple instances of poor ground contact (may be attributed to the low mass and low inertia of the scaled down robot compared to its larger counterpart,  resulting in a narrower friction cone) resulting in foot slippage, and thus decreased performance, especially at higher stride frequencies. 

The $CoT$ measured for HAMR-Jr (\SI{1}{}-\SI{200}{\hertz}) is similar than that of HAMR-VI \cite{goldberg2017gait}\cite{doshi2019effective} and for animals of a similar size \cite{full2010invertebrate}. These results match or are lower than most robots at this scale (Table \ref{tab:hamr_other_comp}). However, the lowest $CoT$ for HAMR-Jr is higher than the same for HAMR-VI matching that observations from biological organisms, which indicate that $CoT$ should increase with decreasing body size \cite{alexander2005models}.

It is also interesting to note that the transmission mechanics show the expected scaling trends (equations \ref{eqn:scale_stiffness} and \ref{eqn:scale_resonance}). However, their values are lower than the predicted estimates (Table \ref{tab:hamr_comp}). We suspect these are likely due to local heat induced variations in mechanical properties of materials resulting from our laser-based fabrication process. While lower lift and swing resonances limit the bandwidth of operational stride frequencies, the decreased vertical stiffness potentially affects the running dynamics. Despite these limitations, we observed a stereotypical aerial phase and leg compression during multiple strides over a range of frequencies (\SI{40}{}-\SI{160}{\hertz}) providing evidence for SLIP-like underlying dynamic behavior \cite{cavagna1977mechanical}. While a detailed analysis of the robot's locomotion dynamics is required to confirm this hypothesis, an estimate of the relative leg stiffness (defined as the ratio of the normalized leg force to the normalized leg compression) of the robot provides a major hint. Relative leg stiffness has been shown to be consistent (\SI{\approx 10}{}) across terrestrial locomotors varying in body size and morphology \cite{blickhan1993similarity} and served as a guiding principle for building dynamically-similar robots. However, due to the allometric scaling applied on HAMR-VI to generate HAMR-Jr, we find that HAMR-Jr (\SI{\approx 63}{}) has a significantly higher relative leg stiffness in contrast to HAMR-VI (\SI{\approx 11}{}) \cite{doshi2019effective}. This suggests that HAMR-Jr is not dynamically similar to HAMR-VI and also marks a departure from a biologically-preferred dynamic regime of locomotion. As immediate next steps, we aim to perform a comprehensive characterization of HAMR-Jr's locomotion dynamics to fully understand the consequence of this allometric scaling.

\begin{table}[th]
\caption{Locomotion performance of HAMR-Jr against other insect-scale robots \cite{st2019toward}}
\label{tab:hamr_other_comp}
\vspace{-5pt}
\begin{center}
\begin{tabular}{l|c|c|c|c}
\hline
\textbf{Robot} & \textbf{Body} & \textbf{Stride} & \textbf{Speed} & \textbf{Cost of}\\
 & \textbf{length} & \textbf{frequency} & \textbf{} & \textbf{Transport}\\
 & \textbf{(\SI{}{\milli\meter})} & \textbf{(\SI{}{\hertz})} & \textbf{(\SI{}{BL\second^{-1}})} & \\
\hline
\hline
 \SI{}Wu (2019) \cite{wu2019insect} & 10 & 850 & 20 & 14\\
  \SI{}St. Pierre (2018) \cite{pierre20183d} & 2.5 & 150 & 14.9 & 33.1\\
 \SI{} HAMR-Jr & 22.5 & 200 & 13.9 & 33.8\\
 \SI{}Shin (2012) \cite{shin2012micro} & 20 & 55-75 & 2.4 & 743\\
 \SI{}Kim (2019) \cite{kim20195} & 2 & 63000 & 4 & -\\
 \SI{}Contreras (2018) \cite{contreras2018six}  & 13 & 100  & 0.07 & 39.8\\
  \SI{}Tribot (2019) \cite{zhakypov2019designing} & 58 & -  & -  & 4.4\\

\hline
\end{tabular}
\end{center}
\end{table}

\section{FUTURE WORK}
\label{sec:future}

Building on the success of HAMR-Jr, we envision a number of exciting research directions for the future. 
Although HAMR-Jr is able to operate at much higher stride frequencies than any terrestrial biological organism \cite{rubin2016exceptional}, it exhibits much lower achievable stride lengths. In the next iteration of this robot, we hope to adjust the transmission ratios (the swing DOF in particular) to realize higher stride lengths and thus increase locomotion speeds. Another cause for reduced running speed was foot slippage. We plan to incorporate miniaturized versions of surface attachment mechanisms previously developed for HAMR \cite{seitz2014bio}. 

Yet another area for improvement of HAMR-Jr is the payload capacity. Under loads greater than its body mass, we observed that the robot's lift DOF was severely limited and prevented the legs from separating from the ground. While the actuators are able to provide sufficient vertical forces, we suspect the off-axis serial compliance (primarily in twist) is limiting the transmission effectiveness. To test this hypothesis, we fabricated a robot with the flexure layer being twice as thick as the current design, as yet another instantiation of a different allometric scaling law in relation to the one described so far in the manuscript. This robot (supplementary video), with no other changes to the design, has a dramatically increased payload capacity and was able to carry at least \SI{3.5}{\gram} (over ten times its body weight). This increased payload capacity is already sufficient to carry onboard, previously developed power and control electronics \cite{goldberg2018power} and sensors \cite{deRivazeaau3038} for successful untethered autonomous operation. As a more permanent solution to this issue, we aim to modify the transmission mechanism to one that limits off-axis rotation better than the current design. Finally, in the long term, we aim to take advantage of our size-agnostic design and fabrication workflow to scale down HAMR-Jr again and determine the limits of this process. We eventually hope to build a suite of highly capable miniature robots at a variety of sizes and work towards approaching the diversity of animal morphologies found in nature.

\addtolength{\textheight}{-8cm}   





\section*{ACKNOWLEDGMENT}

We thank Rut Pe\~{n}a Velasco, Noah Jafferis, and all other members of the Harvard Microrobotics Laboratory for invaluable support and discussions.

\bibliographystyle{IEEEtran}
\bibliography{IEEEabrv,IEEEexample}

\end{document}